# Multi-Objective Optimisation of Cortical Spiking Neural Networks With Genetic Algorithms

James Fitzgerald, and KongFatt Wong-Lin
Intelligent Systems Research Centre, School of Computing, Engineering and Intelligent Systems,
Ulster University, Magee Campus,
Derry~Londonderry, Northern Ireland, UK
{fitzgerald-j5, k.wong-lin}@ulster.ac.uk

*Abstract*— Spiking neural networks (SNNs) communicate through the all-or-none spiking activity of neurons. However, fitting the large number of SNN model parameters to observed neural activity patterns, for example, in biological experiments, remains a challenge. Previous work using genetic algorithm (GA) optimisation on a specific efficient SNN model, using the Izhikevich neuronal model, was limited to a single parameter and objective. This work applied a version of GA, called non-dominated sorting GA (NSGA-III), to demonstrate the feasibility of performing multi-objective optimisation on the same SNN, focusing on searching network connectivity parameters to achieve target firing rates of excitatory and inhibitory neuronal types, including across different network connectivity sparsity. We showed that NSGA-III could readily optimise for various firing rates. Notably, when the excitatory neural firing rates were higher than or equal to that of inhibitory neurons, the errors were small. Moreover, when connectivity sparsity was considered as a parameter to be optimised, the optimal solutions required sparse network connectivity. We also found that for excitatory neural firing rates lower than that of inhibitory neurons, the errors were generally larger. Overall, we have successfully demonstrated the feasibility of implementing multi-objective GA optimisation on network parameters of recurrent and sparse SNN.

## I. Introduction

Spiking neural networks (SNNs) are the third generation of artificial neural networks that attempt to mimic certain functions of real biological brains [1]. SNNs have the ability to encode information based on the spike timing of neurons while using less computational resources than previous generations of neural network models [2]. Spike timing allows the ability for SNNs to encode information in both "space" (i.e., across neurons) and time [3]. SNNs are used not only in computational neuroscience models to understand brain functions, but also used in various applications, including deep learning and knowledge representation, by exploiting its advantage in handling complex temporal or spatiotemporal information [4].

Moreover, with sparsity in network connectivity, as in the biological brain, more complex information can be encoded [5]. Recently, SNNs have been incorporated into large-scale neuromorphic computing systems [6] [7] with the ability to process data and large amounts of information even faster and at lower power. In addition to information coding in spike times, SNNs can also encode information in its firing frequencies (or firing rates), and certain cognitive functions depend on the pooling of the firing rates of the neuronal population [1].

There are various ways to explicitly model spiking neurons, depending on the mechanisms to generate the spiking activity. For example, the simplest type is the perfect integrate-and-fire neuronal model [8]. In consideration of biological realism and computational efficiency, the Izhikevich model is known to generate a wide range of realistic spiking patterns with only a small set of model parameters [9][10], while not being computationally expensive to simulate.

Despite their advantages, the modelling of SNNs poses several challenges. Given the larger number of parameters as compared to classic neural networks, a major challenge is the time-consuming process of model parameter searching [11]. Commonly, optimisation methods are required, which are processes in searching for some optimal solution(s) with respect to the model parameter(s) and some specified goal(s) (via some objective mathematical function(s)) [12]. There are a range of techniques that can be applied depending on the problem space, including finding optimal neural network structures or functions [13]. Popular optimisation techniques include maximum likelihood estimation, gradient descent and genetic algorithm (GA) [14].

In particular, GA, a class of evolutionary algorithms, which is inspired by biological evolution theory, is a heuristic, stochastic, randomised search optimisation technique and it is relatively simple to describe and implement [15] [16] [17]. GA requires no *a priori* knowledge about what it is trying to optimise as domain specific knowledge is contained in the fitness function and the genetic operators defined for the problem. In previous applications of GA to SNNs, [11] proposed a discrete objective function to optimise the size and resilience of SNNs, while [18] used GA to train spiking neural networks to compete for limited resources in simulated environment.

Our previous GA work had shown the feasibility of optimising a single objective function (population-averaged firing rate of specific neuronal type) of an Izhikevich-based recurrently and fully connected SNN model. The model was based on a canonical microcircuit of the mammalian brain [9] [19]. Specifically, as computational neuroscientists or artificial intelligence practitioners often require a certain activity level of the SNN's neuronal population to be set, e.g. to fit to values observed in wet-lab experiments, our previous work optimised the (thalamic) input parameter of the neurons with the objective of minimising the error between the population- and time-averaged firing-rate output and target. However, that work did not explore the possibility of multi-objective optimisation across a set of model parameters including connection/synaptic strengths. Further, the SNN model investigated had biologically unrealistic all-to-all connectivity. Thus, it is unclear whether GA can be used for multi-objective optimisation in a sparsely connected version of the same model.

In this work, the excitatory and inhibitory neurons in the cortical SNN model [9] have their population- and time-averaged neuronal firing rates considered separately as part of a multi-objective function to minimise the distance from their target firing rates. Then, using a version of GA, named nondominated sorting GA (NSGA-III) [20], to optimise the model's connection strengths on excitatory and inhibitory neurons, and the network's connectivity sparsity level. Using the network's connectivity sparsity as part of the multi-objective function was also explored.

## II. SPIKING NEURAL NETWORK MODELLING

### A. Izhikevich neuronal model

The Izhikevich neuronal model is a phenomenological model that mimics biologically realistic spiking patterns without involving the modelling of a variety of ion channel currents [9]. The (trans)membrane potential v of the model can be described by the coupled differential equations [9][10]:

$$\frac{dv}{dt} = 0.04v^2 + 5v + 140 - w + I \quad (1)$$

$$\frac{dw}{dt} = a(bv - w) \quad (2)$$

where $w$ is some recovery variable coupled to $v$, $I$ is the total afferent/input current, and $a$ and $b$ are model parameters that partially determine the spiking characteristics. The model parameter $a$ describes the rate of decay for the neuronal membrane potential $w$ while parameter $b$ represents the sensitivity of the recovery variable to the membrane potential $v$.

When $v$ passes some prescribed peak (set at 30 mV), this results in a neuronal firing or a spike of activity. Upon firing, $v$ is reset to some level $c$ (-65 mV), another model parameter, while the recovery variable $w$ is simultaneously raised by a value of $d$. Both $c$ and $d$ are two additional model parameters, randomised across neurons with values of -65 mV and 2, respectively, which brings to a total of 4 model parameters. Together, these 4 parameters can alter the spiking behavior of a neuron. Unless specified these parameters follow that of [9].

### B. Synapses and network

The SNN model consists of 800 excitatory neurons and 200 inhibitory neurons, based on the observed ratio of excitatory to inhibitory neurons in the mammalian cortex to be around 4:1. The SNN model was implemented by connecting the neurons with excitatory and inhibitory synapses. For simplicity, as in [9][19], this model uses instantaneous current-based synapses [8], i.e., whenever a presynaptic neuron fires a connected postsynaptic neuron will receive an instantaneous pulse-like increase (or decrease) by some value in the postsynaptic current through the term $I$ in Eqn. (1), if the synapses are excitatory (inhibitory).

The range of the synaptic weights can be adjusted with the parameters $ge$ for excitatory neurons and $gi$ for inhibitory neurons. (In previous work [9][19], $ge$ and $gi$ parameters were set constant at values of 0.5 and 1.0, respectively.) The connections between all neurons are represented by some matrix $S$. When investigating network sparsity, some connections in $S$ are set to 0 to mimic sparsity in the network, with fraction of connections, $f$ (= total number of connections in any considered model divided by total possible connections for an all-to-all connectivity model), randomly selected between 0 and 1.

The 4 model parameters to be optimised are mean thalamic input currents, $ge$, $gi$, and $f$. The 2 objective functions are the population- and time-averaged (over 1 second of simulated time) firing rates of the excitatory neurons and inhibitory neurons. Also investigated, was sparsity level as the third objective function, with model parameters thalamic input currents, $ge$ and $gi$, to be optimised.

The original MATLAB code by [9] was rewritten in Python 3.7. Python Pandas [21] software library was used for data manipulation and analysis, while Plotly [22] was used for data visualisation. Fig. 1 shows a sample simulated spike raster diagram (neuron number vs time) of the network model using the same parameters as that in [9].

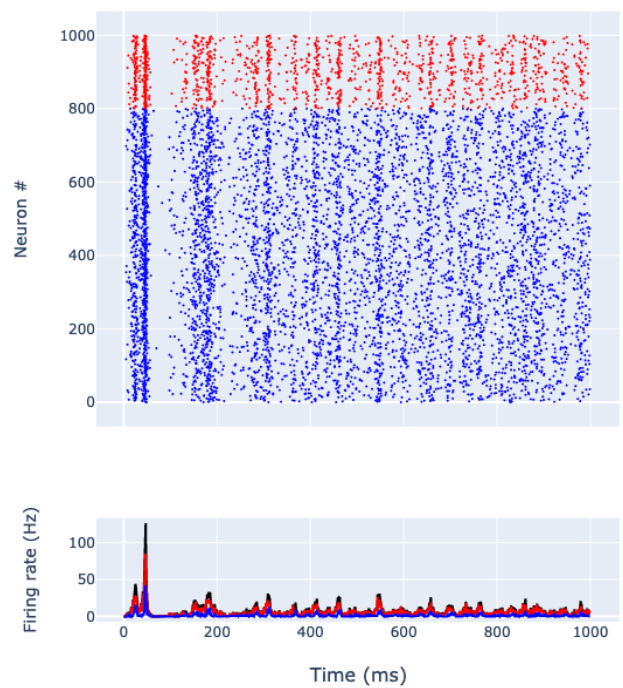

*Fig. 1. Sample simulation of 1000 fully connected spiking neurons over 1000 ms and their aggregated activities. Top: Vertical (horizontal) axis: neuron number (time in ms). 800 excitatory neurons (blue) and 200 inhibitory neurons (red). Each blue/red dot denotes a spike of activity for some neuron at some time point. Model parameters' values based on original model [9]. Bottom: Instantaneous population-averaged firing rates using a time bin of 1 ms. Timescale same as in top panel. Black: All 1000 neurons' averaged firing rates; blue: averaged over 800 excitatory neurons; red: averaged over inhibitory neurons.*

## III. MULTIPLE-OBJECTIVE OPTIMISATION AND GENETIC ALGORITHM

In a single-objective optimisation problem, where some objective function, $G$, is to be maximised (minimised) for some set of $n$ number of parameters, $\{p\}$, the optimal set of parameters is the set that results in maximal (minimal) values of $G$. In multi-objective optimisation e.g. with two objective functions, $G$ and $H$, for some set of parameters $\{p\}$ where one needs to optimise both $G$ and $H$, the complexity of the problem is increased [23]. In particular, if the set of parameter values optimising one objective function is in conflict with another set of parameter values for optimising the other objective function, a trade-off may be required. One such compromising approach is the Pareto frontier or set, which is a set of solutions where $N$ solutions exist where one objective can be made better without making another objective worse [23].

In GAs [15] [16] [17], the procedures typically begin with an initial population which could be randomly generated. Each member of a population is referred to as a chromosome that represents a set of parameters (genes) to reach a solution. A fitness function is applied to every chromosome essentially scoring each solution (chromosome). Informed by the fitness of each chromosome several genetic operators can be applied to the current population to generate the next. Initially the selection operator would be applied, selecting the "fittest" chromosomes to enter the next generation without change. Then a crossover operator is applied selecting pairs of chromosomes to "mate" the likelihood of a chromosome being selected is proportional to its fitness. Finally, a mutation operator is applied to the entire new population, with low probability to make changes to the genes.

A random search for a Pareto set can be performed with minor modifications to the GA described above. Notably, defining an objective function that accepts a vector of continuous or discrete decision variables for ranking. During implementation, the fitness is assumed to be a scalar for each chromosome. This is used to apply genetic operators to the population with respect to the fitness of each solution. However, with multi-objective optimisation there is no single objective to rank the population. This is resolved with some modifications, through introducing a fluctuating population and ranking multiple objectives to create a possible Pareto frontier.

The GA used in this work was the pymoo [24] implementation of nondominated sorting GA (NSGA-III) [25]. A summary of the steps is illustrated in Fig. 2. The GA uses genetic operators to generate new chromosomes for each generation after the entire population has been assessed for fitness of the population. A tournament selection algorithm is applied which randomly selects pairs of chromosomes and pits one against another. The winning ones in the tournament will be identified. Nondominated ones will be selected based on larger crowding distance (i.e. average distance between neighboring solutions) as a secondary criterion. Then the crossover operator is applied to the population using a simulated binary crossover algorithm. Finally, a polynomial mutation is performed to add variance to the set and avoid ending up in a local minimum. See [23] [25] for further details.

The SNN model and NSGA-III algorithm were computed initially using a 1.6 GHz Dual-Core Intel Core i5 with 4 GB 1600 MHz DDR3, and later with Kelvin-2 HPC (8000 AMD-based CPU cores and 32 GPU nodes with a high performance 2 Petabyte of scratch storage interconnected via high-speed network). Source codes for simulating the SNN and NSGA-III are available within Jupiter notebook environments at github.com/FitzgeraldJames/SNNMOO. Data was serialised to csvs and analysed in Jupyter notebook.

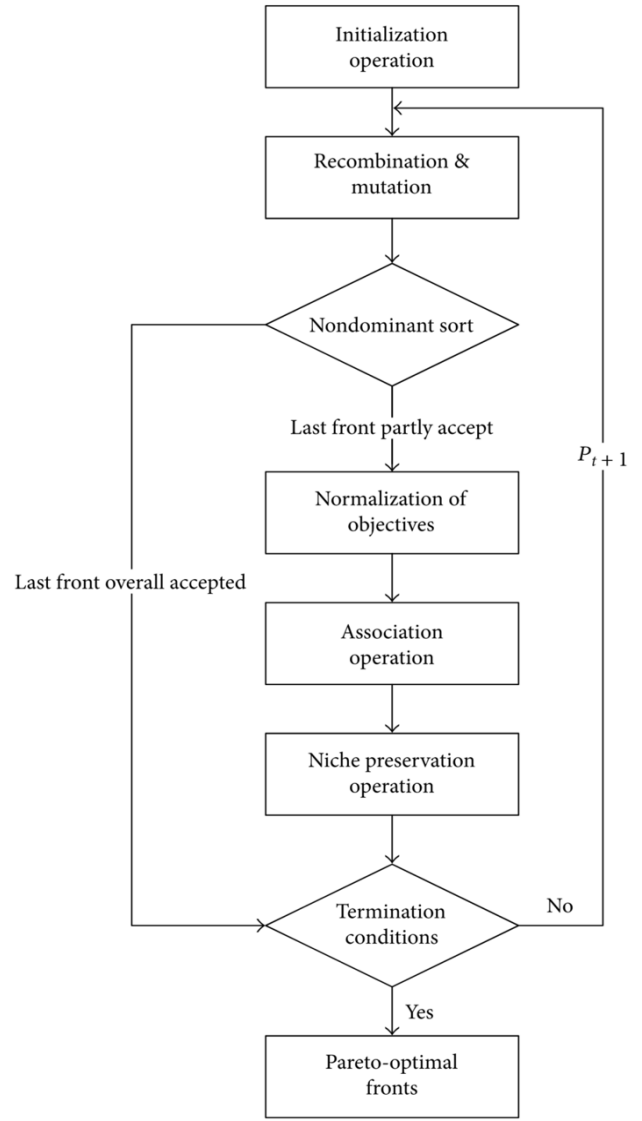

Fig. 2. Flowchart summarising the steps when implementing the nondominated sorting genetic algorithm III (NSGA-III) for multi-objective optimisation problem. Adapted from [27].

IV. MULTIPLE-OBJECTIVE OPTIMISATION OF CORTICAL SNN

As mentioned above, there are two parts to our study. First, for any fixed network connectivity fraction $f$, a search was performed for $ge$ and $gi$ parameters that minimize the (absolute) difference (i.e. the error) between the population- and time-averaged firing rate of the excitatory neurons and inhibitory neurons and their respective targeted values. Second, we include the parameter $f$ together with $ge$ and $gi$ to form a set of parameters to be optimized with respect to the population- and time-averaged firing rates of the excitatory and inhibitory neurons.

For both parts of the study, the Pareto frontiers are mapped out. For the first part of the study, Pareto frontiers could be identified for each value of $f$. Fig. 3 showed the sample Pareto frontiers in the network's output distance from targeted values (of excitatory and inhibitory neural firing rates) while the model parameters, $ge$ and $gi$, were being optimised. The objective targets were 5 Hz and 2 Hz for the excitatory and inhibitory neural firing rates, respectively. Each dot in Fig. 3 represented 25 chromosomes at the end of 50 generations with the NSGA-III algorithm. For this specific target values, the Pareto frontier remained similar despite the fraction of network connectivity $f$ being varied from a value of 1 (all-to-all connectivity case) to 0.2 (legend in Fig. 3).

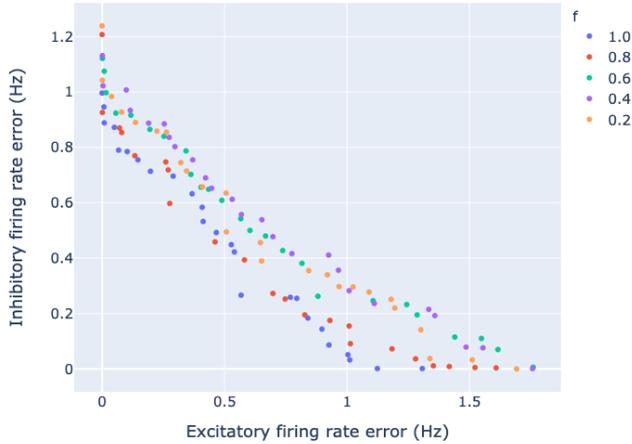

*Fig. 3. Sample Pareto frontiers in network's output errors (of excitatory and inhibitory neural firing rates). Parameters optimized were the range of the synaptic weights to excitatory and inhibitory neurons, ge and gi, respectively. Firing rates were obtained by averaging over all the neurons of the same type during the 1000 ms simulated duration. Dot: 25 chromosomes at the end of 50 generations with the NSGA-III algorithm. Legend and colour label: Different values of fraction of network connectivity, f.*

Next, we investigated different sets of neural firing rate targets for a set of network connectivity fraction values $f$. In. Each $f$ value was optimised for different target excitatory and inhibitory neural firing rate targets (5 Hz, 2 Hz), (2 Hz, 2 Hz), and (5 Hz, 2 Hz). The $f$ values used were 1 (all-to-all connectivity), 0.5 (half the connections), and 0.2 (one fifth the connections).

Fig. 4 illustrated the results for these combinations – a total of 9 combinations. From these results, we observed that for the cases where the targeted excitatory neural firing rate was larger than the targeted inhibitory neural firing rate, the distances between the network outputs and targeted values were generally small (< 2.5 Hz). As the network sparsity increased (smaller $f$ values), the distances slightly increased.

For the cases where the targeted excitatory neural firing rate was smaller than the targeted inhibitory neural firing rate, the distances between the network outputs and targeted values were generally larger (< 20 Hz). The results were not affected by network sparsity. When the targeted excitatory neural firing rate was the same as the targeted inhibitory neural firing rate (at 2 Hz), the distances were intermediate (< 5 Hz). Again, the results were not affected by network sparsity.

Finally, we had the sparsity parameter $f$, together with *ge* and *gi* parameters, to be optimised, with respect to minimising the two neural firing rate distances. Fig. 5 illustrated the Pareto frontiers for three sets of targets, with larger firing rate ranges: (2 Hz, 10 Hz), (5 Hz, 5 Hz) and (10 Hz, 2 Hz). Interestingly, the three targets led to three non-overlapping manifolds of the solutions (Fig. 5, left). Consistent with the results in Fig. 4, we found that the distances (errors) were larger when the excitatory neural firing rates were smaller than that of the inhibitory neurons (Fig. 5, middle, green). When the excitatory neural firing rates were higher than that of inhibitory neurons, the network connectivity had to be very sparse (< 0.16).

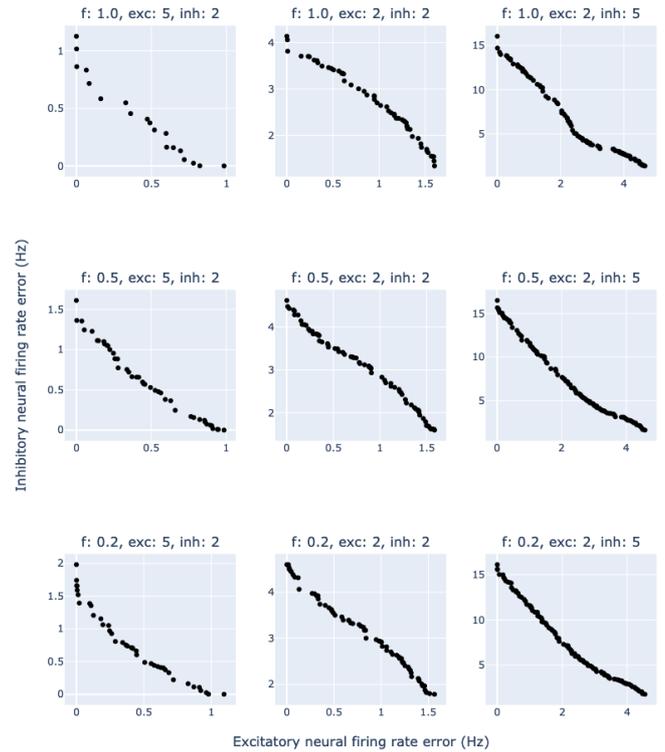

*Fig. 4. Pareto frontiers for different connectivity sparsity, and excitatory and inhibitory neural firing rate targets. Vertical (horizontal) axis label: Distance – difference between network outputs and targeted firing rates. f: fraction of network connectivity; exc: excitatory; inh: inhibitory. Distances in absolute values.*

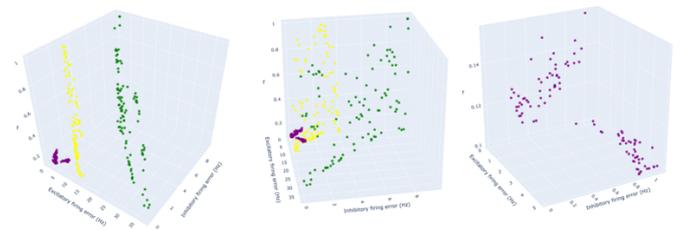

*Fig. 5. Pareto frontiers with network sparsity as a parameter to be optimised. Left: 3D plot for the 3 different sets of target neural firing rates. Green: exc 2 Hz, inh 10 Hz; yellow: exc 5 Hz, inh 5 Hz; purple: exc 10 Hz, inh 2 Hz. Middle: Different angle of view of the same 3D plot in left panel. Right: Zoom in for the target set with excitatory and inhibitory neural firing rates at 10 and 2 Hz, respectively.*

## V. CONCLUSION AND DISCUSSON

In this work, we have successfully applied a version of the GA algorithm, called NSGA-III, in search for optimal model parameter of a cortical column-like recurrent SNN consisting of coupled populations of excitatory and inhibitory neurons. In particular, we were able to search for optimal connectivity parameters with respect to two network firing rate outputs in parallel. The connectivity parameters used were the range of the synaptic weights (*ge* and *gi*), and later the level of connectivity sparsity (*f*, fraction of network connections). We defined the NSGA-III objective function as the distance, i.e. (absolute) difference, between the simulated and targeted

population- and time-averaged firing rates of the excitatory and inhibitory neurons.

Our work was an extension of our previous work [19] in various ways. First, our work utilised a more recent GA algorithm called NSGA-III [20] [25]. Second, unlike our previous work, we optimised for multiple model parameters linked to network connectivity (($ge$, $gi$) or ($ge$, $gi$, $f$)) instead of just a thalamic input current parameter. Third, our work had multiple objectives, the two population firing rates, instead of a single population firing rate as in [19]. Fourth, we made use Pareto frontiers to identify the sets of solutions. Fifth, we investigated how connectivity sparsity (in terms of the fraction of network connections, $f$) influence the optimal sets of solutions.

We found that the NSGA-III algorithm performed the best when the targeted excitatory neural firing rate was larger than the targeted inhibitory neural firing rate. The results were slightly improved with higher connectivity. When the targeted excitatory neural firing rate was smaller than the targeted inhibitory neural firing rate, the algorithm generally had larger distances, and the latter were not affected by network connectivity sparsity. In neurophysiology, under baseline, resting condition in the cortex, regular spiking excitatory pyramidal neurons typically fire below 10 Hz, while fast-spiking inhibitory neurons typically fire within the range of 5-20 Hz [8] [9] [10]. To achieve biological plausibility in which the excitatory neurons fire at a slower rate than that of fast-spiking inhibitory neurons, our work showed that the range of errors could be quite high. This could possibly be due to the not too realistic implementation of instantaneous synapses [8]. In comparison, when excitatory neurons fire at higher rate than that of inhibitory neurons in the model, our work suggested that network connectivity had to be sufficiently sparse.

In this work, GA-based parameter optimization had been performed for only connectivity parameters. With the convenience of the NSGA-III algorithm, it should be relatively straightforward to incorporate other model parameters e.g. intrinsic neuronal model parameters, (thalamic) input current and neuronal noise level. A limitation of our work was the relatively large errors for certain combinations of target firing rates. Future work may require modification of the algorithm, such as additional constraints. In fact, we had conducted further investigations (available in the abovementioned GitHub repository) using additional optimization measures to further reduce the errors in candidate solution space. In particular, by constraining the firing rate solutions, or minimising the distance of candidate solutions from their mean, the algorithms were able to "pull" the Pareto frontier more towards the centre of the error space to further reduce the errors. Future work could also employ similar methods to search for model parameters to fit high-dimensional firing-rate data which is rich in temporal structure (e.g. [26]), and compare performances with other methods, including method using backpropagation through time [27], and explore how such algorithms can be applied to recurrent SNNs in neuromorphic computing system [6].

ACKNOWLEDGMENT

We thank Jose M. Sanchez-Bornot for assisting with our use of the computer cluster simulations. We are grateful for access to the Tier 2 High Performance Computing resources provided by the Northern Ireland High Performance Computing (NI-HPC) facility funded by the UK Engineering and Physical Sciences Research Council (EPSRC), Grant No. EP/T022175/1.